\documentclass[11pt,a4paper]{article}
\usepackage[hyphens]{url}
\usepackage[hyperref]{emnlp2020}
\usepackage{times}
\usepackage{latexsym}

\usepackage{microtype}

\usepackage{booktabs}
\usepackage{comment}
\usepackage{ifthen}
\usepackage{makecell}
\usepackage{multicol}
\usepackage{multirow}
\usepackage{subcaption}
\usepackage{xspace}

\newif\ifcomment
\commenttrue
\usepackage{amsmath}
\usepackage{amssymb}
\usepackage{amsthm}
\usepackage{algorithm}
\usepackage{algorithmic}
\usepackage{graphicx}
\usepackage{accents}
\usepackage{wrapfig}

\theoremstyle{definition}
\providecommand{\corollaryname}{Corollary}

\theoremstyle{plain}
\newtheorem{theorem}{Theorem}[section]
\newtheorem{lemma}{Lemma}[section]
\newtheorem{definition}{Definition}[section]
\newtheorem{proposition}{Proposition}[section]
\newtheorem{assumption}{Assumption}
\newtheorem{corollary}{Corollary}[section]

\newtheoremstyle{TheoremNum}
    {\topsep}{\topsep}              
    {\itshape}                      
    {}                              
    {\bfseries}                     
    {.}                             
    { }                             
    {\thmname{#1}\thmnote{ \bfseries #3}}
\theoremstyle{TheoremNum}

\newcommand\TP{\text{TP}}
\newcommand\FP{\text{FP}}
\newcommand\FN{\text{FN}}

\newcommand\AP{\text{AP}}
\newcommand\hatFP{\hat{\text{FP}}}
\newcommand\wrandom{w_{\text{rand}}}

\newcommand\Din{D_\text{statedneg}}
\newcommand\Dpos{D_\text{pos}}

\newcommand\Dtrainpairs{D^\text{train}_\text{all}}
\newcommand\Dclose{D^\text{train}_\text{close}}
\newcommand\Dtrainin{D^\text{train}_\text{heur}}

\newcommand\Ddevpairs{D^\text{dev}_\text{all}}
\newcommand\Ddevnear{D^\text{dev}_\text{near}}
\newcommand\Ddevrandom{D^\text{dev}_\text{rand}}

\newcommand\Dtestpairs{D^\text{test}_\text{all}}
\newcommand\Dtestnear{D^\text{test}_\text{near}}
\newcommand\Dtestrandom{D^\text{test}_\text{rand}}
\newcommand\Dtestneg{D^\text{test}_\text{neg}}
\newcommand\Dtestin{D^\text{test}_\text{heur}}

\newcommand{\nl}[1]{``\textit{#1}''}
\newcommand{\vanillabert}{\textsc{ConcatBERT}\xspace}
\newcommand{\ourmodel}{\textsc{CosineBERT}\xspace}

\newcommand\sX{\ensuremath{\mathcal{X}}}

\newcommand\R{\ensuremath{\mathbb{R}}} 

\newcommand\refsec[1]{Section~\ref{sec:#1}}

\newcommand\reffig[1]{Figure~\ref{fig:#1}}

\newcommand\reftab[1]{Table~\ref{tab:#1}}
\newcommand\refapp[1]{Appendix~\ref{sec:#1}}

\ifthenelse{\isundefined{\definition}}{\newtheorem{definition}{Definition}}{}
\ifthenelse{\isundefined{\assumption}}{\newtheorem{assumption}{Assumption}}{}
\ifthenelse{\isundefined{\hypothesis}}{\newtheorem{hypothesis}{Hypothesis}}{}
\ifthenelse{\isundefined{\proposition}}{\newtheorem{proposition}{Proposition}}{}
\ifthenelse{\isundefined{\theorem}}{\newtheorem{theorem}{Theorem}}{}
\ifthenelse{\isundefined{\lemma}}{\newtheorem{lemma}{Lemma}}{}
\ifthenelse{\isundefined{\corollary}}{\newtheorem{corollary}{Corollary}}{}
\ifthenelse{\isundefined{\alg}}{\newtheorem{alg}{Algorithm}}{}
\ifthenelse{\isundefined{\example}}{\newtheorem{example}{Example}}{}

\aclfinalcopy 

\title{On the Importance of Adaptive Data Collection for \\ Extremely Imbalanced Pairwise Tasks}
\author{Stephen Mussmann\thanks{\ Authors contributed equally.} \qquad
  Robin Jia\footnotemark[1] \qquad 
  Percy Liang \\
  Computer Science Department, Stanford University, Stanford, CA \\
  \texttt{mussmann@stanford.edu} \qquad \texttt{\{robinjia,pliang\}@cs.stanford.edu} 
  }

\date{}

\begin{document}
\maketitle
\begin{abstract}
Many pairwise classification tasks, such as paraphrase detection and open-domain question answering, naturally have extreme label imbalance (e.g., $99.99\%$ of examples are negatives).
In contrast, many recent datasets heuristically choose examples to ensure label balance.
We show that these heuristics lead to trained models that generalize poorly:
State-of-the art models trained on QQP and WikiQA each have only $2.4\%$ average precision 
when evaluated on realistically imbalanced test data.
We instead collect training data with active learning,
using a BERT-based embedding model to efficiently retrieve uncertain points from a very large pool of unlabeled utterance pairs.
By creating balanced training data with more informative negative examples, active learning greatly improves average precision to $32.5\%$ on QQP and $20.1\%$ on WikiQA.
\end{abstract}

\section{Introduction}

For most pairwise classification tasks in NLP, the most realistic data distribution has extreme label imbalance (e.g., $99.99\%$ of examples have the same label).
In question deduplication \citep{iyer2017qqp}, the vast majority of pairs of questions from an online forum are not duplicates. In open-domain question answering \citep{yang2015wikiqa,lee2019latent}, almost any randomly sampled document will not answer a given question.
Random pairs of sentences from a diverse distribution will have no relation between them in natural language inference \citep{bowman2015large}, as opposed to an entailment or contradiction relationship.

While past work has recognized the importance of label imbalance in NLP \citep{lewis2004rcv1,chawla2004imbalanced},
many recently released datasets are heuristically collected to ensure label balance, generally for ease of training. For instance, the Quora Question Pairs (QQP) dataset \citep{iyer2017qqp} was generated by mining non-duplicate questions that were heuristically determined to be near-duplicates. The SNLI dataset had crowdworkers generate inputs to match a specified label distribution \citep{bowman2015large}. In this work, we show that models trained on heuristically balanced datasets 
deal poorly with natural label imbalance at test time:
They have very low average precision on realistically imbalanced test data created by taking all pairs of test utterances.

\begin{figure}
  \centering
  \includegraphics[width=\linewidth]{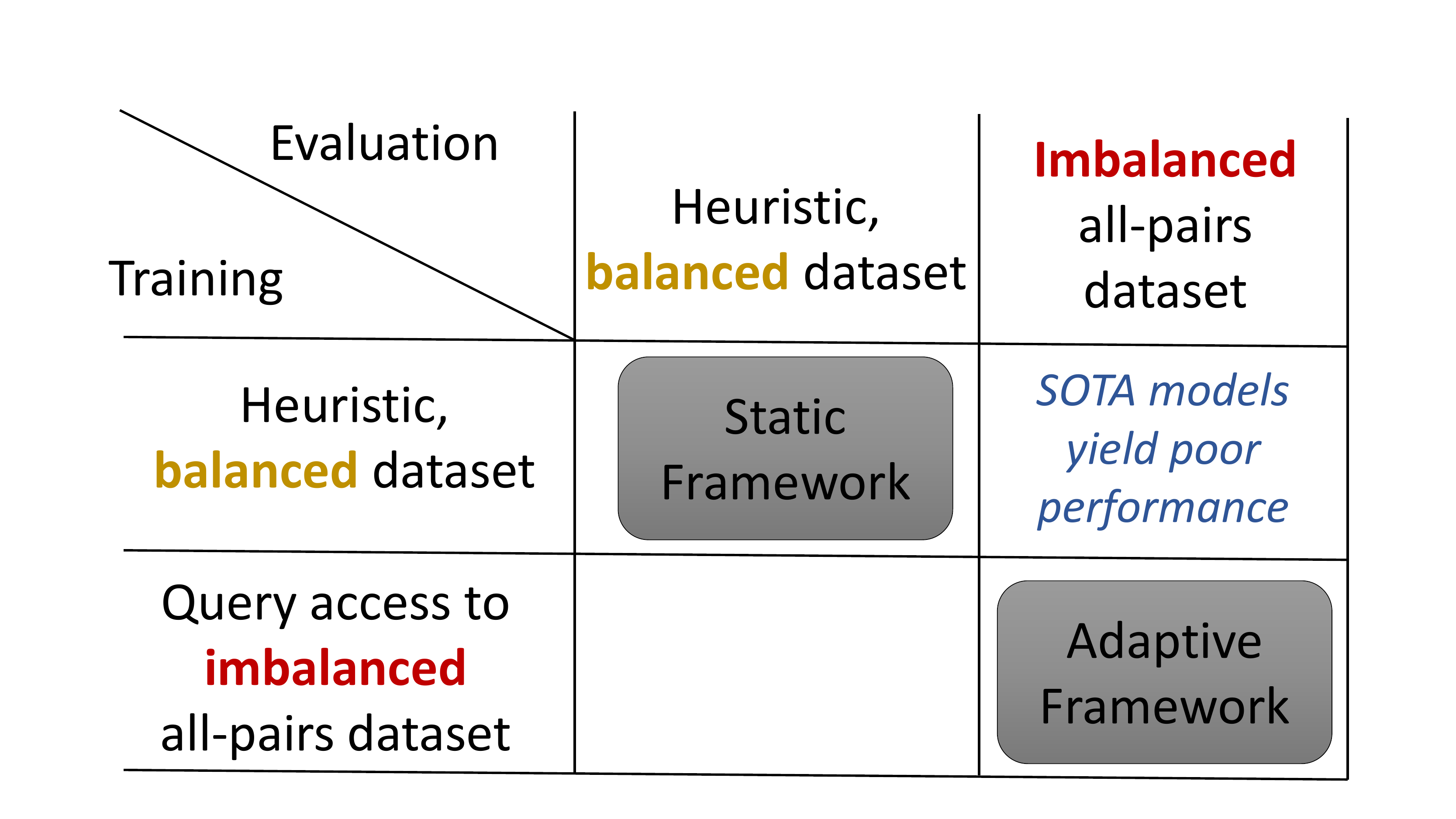}
  \caption{
  Modern benchmarks often use heuristically balanced data for training and evaluation.
  We find that models trained on this data perform poorly on the very imbalanced all-pairs distribution and develop adaptive methods to collect training data for this setting.
  }
  \label{fig:schamatic-diagram}
\end{figure}

Instead of heuristically producing static training datasets, we study adaptive data collection methods.
In particular, we apply the pool-based active learning framework \citep{settles2009active} to extremely imbalanced pairwise tasks. At training time, the system has \emph{query access} (i.e., the ability to collect a limited subset of the labels) to a large unlabeled dataset formed by taking all utterance pairs from a set of training utterances.
For example, in the question deduplication setting, we might have a budget to annotate pairs of questions as ``duplicate'' or ``not duplicate,'' and wish to train a model on this data to find new duplicate pairs with high precision.

Data collection for extremely imbalanced pairwise tasks is challenging because the pool of unlabeled examples is very large (as it grows quadratically in the number of utterances)
and very few of the examples are positive.
We collect balanced training data using uncertainty sampling, an adaptive method that queries labels for examples on which a model trained on previously queried data has high uncertainty \citep{lewis1994sequential}.
To lower the computational cost of searching for uncertain points,
we propose combining active learning with a BERT embedding model for which uncertain points can be located efficiently using nearest neighbor search.

In this work, we empirically show that our use of adaptive data collection yields significant gains over static heuristics. In order to compare methods without collecting data separately for each method and each run, we perform retrospective data collection with imputed labels to simulate data collection. Uncertainty sampling with our BERT embedding model achieves $32.5\%$ and $20.1\%$ average precision for QQP and WikiQA, respectively. In contrast, state-of-the-art models trained on the original heuristically collected data each have only average precision of $2.4\%$.
\section{Setting}

The pairwise tasks described above fall under a more general category of binary classification tasks, those with an input space $\sX$ and output space $\{0,1\}$. 
We assume the label $y$ is a deterministic function of $x$, which we write $y(x)$. A classification model $p_\theta$ yields probability estimates $p_\theta(y \mid x)$ where $x \in \sX$.

Our setting has two aspects: The way training data is collected via label queries (\refsec{framework-data}) and the way we evaluate the model $p_\theta(y \mid x)$ by measuring average precision (\refsec{framework-eval}). 
This work focuses on pairwise tasks (\refsec{pairwise}), which enables efficient active learning (\refsec{active}).

\subsection{Data collection}
\label{sec:framework-data}
In our setting, a system is given an unlabeled dataset $\Dtrainpairs \subseteq \sX$.
The system can query an input $x \in \Dtrainpairs$ and receive the corresponding label $y(x)$. 
The system is given a budget of $n$ queries to build a labeled training dataset of size $n$.

\subsection{Evaluation}
\label{sec:framework-eval}
Following standard practice for imbalanced tasks,
we evaluate on precision, recall, and average precision \citep{lewis1995evaluating,manning2008ir}.
A scoring function $S: \sX \rightarrow \R$ (e.g., $S(x) = p_\theta(y=1 \mid x)$)
is used to rank examples $x$, where an ideal $S$ assigns all positive examples $\{x: y(x) = 1\}$ a higher score than all negative examples $ \{x: y(x) = 0\}$.
Given a test dataset $\Dtestpairs \subseteq \sX$, define the number of true positives, false positives, and false negatives of a scoring function $S$ at a threshold $\gamma$ as:

\begin{align}
    \TP(S,\gamma) &= \sum_{x \in \Dtestpairs} \textbf{1}[ y(x)=1  \wedge S(x) \geq \gamma ] \\
    \FP(S,\gamma) &= \sum_{x \in \Dtestpairs} \textbf{1}[ y(x)=0  \wedge S(x) \geq \gamma ] \\
    \FN(S,\gamma) &= \sum_{x \in \Dtestpairs} \textbf{1}[ y(x)=1  \wedge S(x) < \gamma ].
\end{align}

For any threshold $\gamma$, define the precision $P(S, \gamma)$ and recall $R(S, \gamma)$ of a scoring function $S$ as:
\begin{align}
    \text{P}(S,\gamma) &= \frac{\TP(S,\gamma)}{\TP(S,\gamma) + \FP(S,\gamma)} \\
    \text{R}(S,\gamma) &= \frac{\TP(S,\gamma)}{\TP(S,\gamma) + \FN(S,\gamma)}.
\end{align}

Let $\Gamma = \{S(x): x\in \Dtestpairs\}$ be the set of scores of the dataset.
By sweeping over all distinct values $\Gamma$ in descending order, we trace out the precision-recall curve. The area under the precision recall curve or average precision (AP) is defined as:
\begin{align}
    \AP(S) = \sum_{i=1}^{|\Gamma|} (\text{R}(S,\gamma_i) - \text{R}(S,\gamma_{i-1}) ) \text{P}(S,\gamma_i),
\end{align}
where $\gamma_0 = \infty$ and $\gamma_1 > \gamma_2 > \dots \gamma_{|\Gamma|}$ and $\gamma_i \in \Gamma$.

Note that high precision requires very high accuracy when the task is extremely imbalanced. 
For example, if only one in $10,000$ examples in $\Dtestpairs$ is positive and 50\% precision at some recall is achieved, that implies 99.99\% accuracy.

\subsection{Pairwise tasks}
\label{sec:pairwise}

In this work, we focus on ``pairwise'' tasks, meaning that the input space decomposes as $\sX = \sX_1 \times \sX_2$. For instance, $\sX_1$ could be questions and $\sX_2$ could be paragraphs for open-domain question answering, and $\sX_1=\sX_2$ could be questions for question deduplication. We create the unlabeled \emph{all-pairs} training set $\Dtrainpairs$ by taking the cross product of a subset from $\sX_1$ and a subset from $\sX_2$. We follow the same procedure to form the all-pairs test set, $\Dtestpairs$. As is standard practice, we ensure that the train and test all-pairs sets are disjoint.

Many pairwise tasks require high average precision on all-pairs test data.
A question deduplication system must compare a new question with all previously asked questions to determine if a duplicate exists.
An open-domain question-answering system must search through all available documents for one that answers the question.
In both cases, the all-pairs distribution is extremely imbalanced, as the vast majority of pairs are negatives, while standard datasets are artificially balanced.
\section{Results training on heuristic datasets}
In this section, we show that state-of-the-art models trained on
two standard pairwise classification datasets---QQP and WikiQA---do not generalize well to our extremely imbalanced all-pairs test data, which we create from an original dataset by imputing a negative label for all pairs that are not marked as positive.
Both QQP and WikiQA were collected using static heuristics 
that attempt to find points $x \in \sX$ that are more likely to be positive.
These heuristics are necessary because uniformly sampling from $\sX$ is impractical due to the label imbalance:
if the proportion of positives is $10^{-4}$, 
then random sampling would have to label 10,000 examples on average to find one positive example.
Standard models can achieve high test accuracy on test data collected with these heuristics, 
but fare poorly when evaluated on all-pairs data derived from the same data source  (\refsec{std-eval}).
Manual inspection confirms that these models often make surprising false positive errors (\refsec{manual}).

\subsection{Experimental setup}
\subsubsection{Evaluation}
We evaluate models on both heuristically balanced test data and our imbalanced all-pairs test data. 

\paragraph{Heuristically balanced evaluation.}
Let $\Dpos$ denote the set of all positive examples,
and $\Din$ denote the set of \emph{stated negative examples}---negative examples in the original heuristically collected dataset.
We define the stated test dataset $\Dtestin$ as the pairs in the original balanced dataset that are also in our defined test dataset: $(\Dpos \cup \Din) \cap \Dtestpairs$.
This is similar to the original QQP test data, but with a different train/test split.
We use task-specific evaluation metrics described in the next section.

\paragraph{All-pairs evaluation.}
All-pairs evaluation metrics depend on the label of every pair in $\Dtestpairs$. 
We approximate these labels by imputing (possibly noisy) labels on all pairs 
using the available labeled data, as described in the next section.\footnote{
For settings where labels cannot be imputed reliably, precision can be estimated by labeling predicted positives,
and recall can be estimated with respect to a non-exhaustive set of known positives
\citep{harman1992overview,ji2011kbp}.
} 
In \refsec{manual}, we manually label examples to confirm our results from this automatic evaluation.

Computing the number of false positives $\FP(S, \gamma)$ requires enumerating all negative examples
in $\Dtestpairs$, which is too computationally expensive with our datasets.
To get an unbiased estimate of $\FP(S, \gamma)$, we could randomly subsample $\Dtestpairs$,
but the resulting estimator has high variance.
We instead compute an unbiased estimator that uses importance sampling.
In particular, we combine counts of errors on a set of ``nearby negative'' examples $\Dtestnear \subseteq \Dtestpairs$, pairs of similar utterances on which we expect more false positives to occur,
and random negatives $\Dtestrandom$ sampled uniformly from negatives in $\Dtestpairs \setminus \Dtestnear$.
Details are provided in \refapp{app-evaluation}.

\subsubsection{Datasets}
\paragraph{Quora Question Pairs (QQP).} 
The task for QQP \citep{iyer2017qqp} is to determine whether two questions are paraphrases.
The non-paraphrase pairs in the dataset were chosen heuristically, e.g., by finding questions on similar topics.
We impute labels on all question pairs by assuming that two questions are paraphrases if and only if they are equivalent under the transitive closure of the equivalence relation defined by the labeled paraphrase pairs.\footnote{Using the transitive closure increases the total number of positives from 149,263 to 228,548, so this adds many positives but does not overwhelm the original data.}
We randomly partition all unique questions into 
train, dev, and test splits,
ensuring that no two questions that were paired (either in positive or negative examples) in the original dataset end up in different splits. 
Since every question is a paraphrase of itself, we define $\Dtrainpairs$ as the set of \emph{distinct} pairs of questions from the training questions, and define $\Ddevpairs$ and $\Dtestpairs$ analogously.
For heuristically balanced evaluation, we report accuracy and F1 score on $\Dtestin$, as in \citet{wang2019glue}.

\paragraph{WikiQA.} 
The task for WikiQA \citep{yang2015wikiqa} is to determine whether a question is answered by a given sentence.
The dataset only includes examples that pair a question with sentences from a Wikipedia article believed to be relevant based on click logs.
We impute labels by assuming that question-sentence pairs not labeled in the dataset are negative.
We partition the questions into
train, dev, and test,
following the original question-based split of the dataset,
and then take the direct product with the set of all sentences in the original dataset to form the train, dev, and test sets.
For all WikiQA models, we prepend the title of the source article to the sentence to give the model information about the sentence's origin, as in \citet{lee2019latent}.

For heuristically balanced evaluation, we report two evaluation metrics.
Following standard practice, we report clean mean average precision (c-MAP), defined as MAP over ``clean'' test questions---questions that are involved in both a positive and negative example in $\Dtestin$ 
\citep{garg2020tanda}.
We also report F1 score across all examples in $\Dtestin$ (a-F1), which unlike c-MAP 
considers the more realistic setting where questions may not be answerable given the available article.
This introduces more label imbalance, as positives make up $6\%$ of $\Dtestin$
but $12\%$ of clean examples.
The original WikiQA paper advocated a-F1 \citep{yang2015wikiqa},
but most subsequent papers do not report it
\citep{shen2017inter,yoon2019compare,garg2020tanda}.

\begin{table}[t]
\centering
\fontsize{8}{9}\selectfont
\begin{tabular}{lccccc}
\toprule
Split & Positives & \makecell{Total \\ pairs} & Ratio & \makecell{Stated \\ Neg.} & \makecell{Nearby \\ Neg.} \\
\midrule
\textbf{QQP} & & & & \\
Train    & 124,625 & 38B  & 1:300K & 132,796 & - \\
Dev      & 60,510  & 8.8B & 1:150K & 61,645 & 13.1M \\
Test     & 43,413  & 8.5B & 1:190K & 60,575 & 12.8M\\
\midrule
\textbf{WikiQA} & & & & \\
Train & 1,040   & 56M  & 1:53K & 19,320 & - \\
Dev   & 140     & 7.8M & 1:56K & 2,593  & 29,511 \\
Test  & 293     & 17M  & 1:57K & 5,872  & 63,136 \\
\bottomrule
\end{tabular}
\caption{Statistics of our QQP and WikiQA splits.}
\label{tab:data}
\end{table}

\paragraph{Data statistics.}
\reftab{data} shows dataset statistics.
Models in this section are trained on the \emph{stated training dataset}
$\Dtrainin \triangleq (\Dpos \cup \Din) \cap \Dtrainpairs$,
the set of all positives and heuristic negatives in the train split.
For all-pairs evaluation, both QQP and WikiQA have extreme label imbalance: Positive examples make up between 1 in 50,000 (WikiQA) and 1 in 200,000 (QQP) of the test examples.

\subsubsection{Models}%
We train four state-of-the-art models that use BERT-base \citep{devlin2019bert}, XLNet-base \citep{yang2019xlnet}, RoBERTa-base \citep{liu2019roberta}, and ALBERT-base-v2 \citep{lan2020albert}, respectively.
As is standard, all models receive as input the concatenation of $x_1$ and $x_2$ separated by a special token;
we refer to these as concatenation-based (\textsc{Concat}) models.
We train on binary cross-entropy loss for 2 epochs on QQP and 3 epochs on WikiQA, chosen to maximize dev all-pairs AP for RoBERTa. 
We report the average over three random seeds for training.

\begin{table}[t]
\centering
\small
\begin{tabular}{l|cc|cc}
\toprule
\multirow{2}{*}{QQP} & \multicolumn{2}{c|}{\textbf{Heur. Balanced}} & \multicolumn{2}{c}{\textbf{All pairs}} \\
& Accuracy & F1 & P@R20 & AP \\
\midrule
BERT    & $82.5\%$ & $77.3\%$ & $3.0\%$ & $2.4\%$ \\
XLNet   & $83.0\%$ & $77.9\%$ & $1.7\%$ & $1.4\%$ \\
RoBERTa & $84.4\%$ & $80.2\%$ & $2.5\%$ & $2.0\%$ \\
ALBERT  & $79.6\%$ & $73.0\%$ & $3.5\%$ & $1.9\%$ \\
\bottomrule
\toprule
WikiQA & c-MAP & a-F1 & P@R=20 & AP \\
\midrule
BERT    & $79.9\%$ & $45.9\%$ & $6.5\%$ & $2.4\%$ \\
XLNet   & $80.5\%$ & $46.7\%$ & $1.0\%$ & $1.0\%$ \\
RoBERTa & $84.6\%$ & $53.6\%$ & $3.4\%$ & $2.3\%$ \\
ALBERT  & $78.2\%$ & $41.8\%$ & $0.7\%$ & $0.9\%$ \\
\bottomrule
\end{tabular}
\caption{State-of-the-art \textsc{Concat} models trained on heuristically collected data
generalize to test data from the same distribution, 
but not to all-pairs data.} 
\label{tab:std-eval}
\end{table}

\subsection{Evaluation results}
\label{sec:std-eval}
As shown in \reftab{std-eval}, state-of-the-art models trained only on stated training data do well on heuristically balanced test data but poorly on the extremely imbalanced all-pairs test data.
On QQP, the best model gets $80.2\%$ F1 on heuristically balanced test examples.\footnote{
On the GLUE QQP train/dev split, our RoBERTa implementation gets $91.5\%$ dev accuracy.
Our in-domain accuracy numbers are lower due to our more challenging train/test split,
as discussed in \refapp{app-glue}.}
However, on all-pairs test data,
the best model can only reach $3.5\%$ precision at a modest $20\%$ recall.
On WikiQA, our best c-MAP of $84.6\%$ is higher than the best previously reported c-MAP without using additional question-answering data, $83.6\%$ \citep{garg2020tanda}.
However, on all-pairs test data, the best model gets $6.5\%$ precision at $20\%$ recall.
All-questions F1 on heuristically balanced data is also quite low, with the best model only achieving $53.6\%$.
Since a-F1 evaluates on a more imbalanced distribution than c-MAP, this further demonstrates that state-of-the-art models deal poorly with test-time label imbalance.
Compared with a-F1, all-pairs evaluation additionally shows that models make many mistakes when evaluated on questions paired with less related sentences;
these examples should be easier to identify as negatives, but are missing from $\Dtrainin$.

\begin{figure}[t]
\begin{center}

\textbf{QQP, \vanillabert trained on $\Dtrainin$} \\
\vspace{0.1in}
\noindent
{\small
$x_1$: \nl{How do I overcome \textbf{seasonal affective disorder}?} \\
$x_2: $\nl{How do I solve \textbf{puberty problem}?} \\
}

\vspace{-0.1in}
\rule{0.25\textwidth}{.2pt}

{\small
$x_1$: \nl{What will \textbf{10000} A.D be like?} \\
$x_2: $\nl{Does not introduction of new \textbf{Rs.2000} notes ease carrying black money in future?} \\
}
\vspace{-0.1in}
\rule{0.25\textwidth}{.2pt}

{\small
$x_1$: \nl{Can a person with no Coding knowledge learn \textbf{Machine learning}?} \\
$x_2$: \nl{How do I learn \textbf{Natural Language Processing}?} \\
}

\vspace{0.1in} \hrule \vspace{0.1in}

\textbf{WikiQA, \vanillabert trained on $\Dtrainin$} \\
\vspace{0.1in}

{\small
$x_1$: \nl{where does \textbf{limestone} form?} \\ 
$x_2$: \nl{Glacier cave . A \textbf{glacier cave} is a cave formed within the ice of a glacier .} \\
}

\vspace{-0.1in}
\rule{0.25\textwidth}{.2pt}

{\small
$x_1$: \nl{what is \textbf{gravy} made of?} \\
$x_2$: \nl{\textbf{Amaretto}. It is made from a base of apricot pits or almonds, sometimes both.} \\
}

\vspace{0.1in} \hrule \vspace{0.1in}

\caption{Examples of confident false positives from the all-pairs test distribution for models trained on examples from the original QQP and WikiQA datasets. Bold highlights non-equivalent phrases.
}
\label{fig:blind-spots}
\end{center}
\end{figure}

\subsection{Manual verification of imputed negatives}
\label{sec:manual}
Our all-pairs evaluation results are based on automatically imputed negative labels, rather than the gold label evaluation metrics. To check the validity of our results, we manually labeled putative false positive errors---examples that our model labeled positively but for which the imputed label was negative---to  more accurately estimate precision.
We focused on the best QQP model and random seed combination on the development set, 
which got $8.2\%$ precision at $20\%$ recall.\footnote{By manual inspection, QQP had more borderline cases than WikiQA, so we focused on QQP.}
For this recall threshold, we manually labeled 50 randomly chosen putative false positives from $\Ddevnear$, and 50 more from $\Ddevrandom$.
In $72\%$ and $92\%$ of cases, respectively, the imputed label was correct and the model was wrong.
Extrapolating from these results, we estimate the true precision of the model to be $9.5\%$, still close to our original estimate of $8.2\%$.
See \refapp{app-manual-update} for more details. For the remainder of the paper, we simply use the imputed labels, keeping in mind this may underestimate precision.

\reffig{blind-spots} shows real false positive predictions at $20\%$ recall for the best QQP and WikiQA models.
For QQP, models often make surprising errors on pairs of unrelated questions (first two examples),
as well as questions that are somewhat related but distinct (third example).
For WikiQA, models often predict a positive label when something in the sentence has the same type as the answer to the question, even if the sentence and question are unrelated.
While these pairs seem easy to classify, the heuristically collected training data lacks coverage of these pairs, leading to poor generalization.
\section{Active learning for pairwise tasks}
\label{sec:active}

As shown above, training on heuristically collected balanced data leads to low average precision on all pairs.
How can we collect training data that leads to high average precision? 
We turn to active learning, in which new data is chosen adaptively based on previously collected data.
Adaptivity allows us to ignore the vast majority of obvious negatives (unlike random sampling)
and iteratively correct the errors of our model (unlike static data collection) by collecting more data around the model's decision boundary.

\subsection{Active learning}

Formally, an active learning method 
takes in an unlabeled dataset $\Dtrainpairs \subseteq \sX$. Data is collected in a series of $k>1$ rounds. 
For the $i^{th}$ round, we choose a batch $B_i \subseteq \Dtrainpairs$ of size $n_i$ and observe the outcome as the labels $\{(x, y(x)): x \in B_i\}$. The budget $n$ is the total number of points labeled, i.e., $n = \sum_{i=1}^k n_i$.
This process is adaptive because we can choose batch $B_i$ based on the labels of the previous $i-1$ batches.
Static data collection corresponds to setting $k=1$.

\paragraph{Uncertainty sampling.}
The main active learning algorithm we use is uncertainty sampling \citep{lewis1994sequential}, which is simple, effective, and commonly used in practice \citep{settles2009active}.
Uncertainty sampling first uses a static data collection method to select the \emph{seed set} $B_1$.
For the next $k-1$ rounds,
uncertainty sampling trains a model on all collected data
and chooses $B_i$ to be the $n_i$ unlabeled points in $\Dtrainpairs$ 
on which the model is most uncertain.
For binary classification,
the most uncertain points are the points where $p_\theta(y=1 \mid x)$ is closest to $\frac12$.
Note that a brute force approach to finding $B_i$ requires evaluating $p_\theta$ on every example in $\Dtrainpairs$, which can be prohibitively expensive.
In balanced settings, it suffices to choose the most uncertain point from a small random subset of $\Dtrainpairs$ \citep{ertekin2007learning};
however, this strategy works poorly in extremely imbalanced settings, as a small random subset of $\Dtrainpairs$ is unlikely to contain any uncertain points.
In \refsec{implementation}, we address this computational challenge with a bespoke model architecture.

\paragraph{Adaptive retrieval.}
We also use a related algorithm we call adaptive retrieval,
which is like uncertainty sampling but queries
the $n_i$ unlabeled points in $\Dtrainpairs$ with highest $p_\theta(y=1 \mid x)$ (i.e., pairs the model is most confident are positive).
Adaptive retrieval can be seen as greedily maximizing the number of positive examples collected.

\subsection{Modeling and implementation}
\label{sec:implementation}
We now fully specify our approach by describing
our model, how we find pairs in the unlabeled pool $\Dtrainpairs$ to query,
and how we choose the seed set $B_1$.
In particular, a key technical challenge is that the set of training pairs $\Dtrainpairs$ is too large to enumerate, as it grows quadratically.
We therefore require an efficient way to locate the uncertain points in $\Dtrainpairs$.
We solve this problem with a model architecture \ourmodel that enables efficient nearest neighbor search \citep{gillick2019learning}.

\subsubsection{Model}
\label{sec:model}
Given input $x = (x_1, x_2)$, \ourmodel embeds $x_1$ and $x_2$ independently and predicts $p_\theta(y=1 \mid x)$ based on vector-space similarity.
More precisely,
\begin{align}
    p_\theta(y=1 \mid x) = \sigma \left(w \cdot \frac{e_\theta(x_1) \cdot e_\theta(x_2)}{\|e_\theta(x_1)\| \|e_\theta(x_2)\|} + b \right),
\end{align}
where $\sigma$ is the sigmoid function, $w > 0$ and $b$ are learnable parameters, and $e_\theta: \sX_1 \cup \sX_2 \rightarrow \R^d$ is a learnable embedding function.
In other words, we compute the cosine similarity of the embeddings of $x_1$ and $x_2$, and predict $y$ using a logistic regression model with cosine similarity as its only feature.
We define $e_\theta$ as the final layer output of a BERT model \citep{devlin2019bert} mean-pooled across all tokens \citep{reimers2019sentence}.\footnote{Although WikiQA 
involves an asymmetric relationship between questions and sentences, we use the same encoder for both.
This is still expressive enough for WikiQA, 
since the set of questions and set of sentences are disjoint.
For asymmetric tasks like NLI where $\sX_1 = \sX_2$, 
we would need to use separate encoders for the $\sX_1$ and $\sX_2$.}
\citet{gillick2019learning} used a similar model for entity linking.

\subsubsection{Finding points to query}
\label{sec:finding}
Next, we show how to choose the batch $B_i$ of points to query,
given a model $p_\theta(y \mid x)$ trained on data from batches $B_1, \dotsc, B_{i-1}$.
Recall that uncertainty sampling chooses the points $x$ for which
for which $p_\theta(y=1 \mid x)$ is closest to $\frac12$,
and adaptive retrieval chooses the points $x$ with largest $p_\theta(y=1\mid x)$.
Since the set of positives is very small compared to the size of $\Dtrainpairs$, 
the set of uncertain points can be found by finding points with the largest $p_\theta(y=1 \mid x)$, thus filtering out the confident negatives, and then selecting the most uncertain from those.

To find points with largest $p_\theta(y=1 \mid x)$, we leverage the structure of our model.
Since $w>0$, $p_\theta(y=1 \mid x)$ is increasing in the cosine similarity of $e_\theta(x_1)$ and $e_\theta(x_2)$. 
Therefore, it suffices to find pairs $(x_1, x_2)$ that are nearest neighbors in the embedding space defined by $e_\theta$.
In particular, for each $x_1 \in \sX_1$, we use the Faiss library 
\citep{JDH17} to retrieve a set $N(x_1)$ containing the $m$ nearest neighbors in $\sX_2$, and define $\Dclose$ to be the set of all pairs $(x_1, x_2)$ such that $x_2 \in N(x_1)$.
We then iterate through $\Dclose$
to find either the most uncertain points (for uncertainty sampling) or points with highest cosine similarity (for adaptive retrieval).
Note that this method only requires embedding the number of distinct elements that appear in the training set, rather than the total number of pairs, the requirement for jointly embedding all pairs.

\subsubsection{Choosing the seed set}
\label{sec:seed-set}
Recall that both of our active learning techniques require a somewhat representative initial seed set $B_1$ to start the process. 
We use the pre-trained BERT model as the embedding $e_\theta$ and select the $n_1$ pairs with largest $p_\theta(y=1 \mid x)$. Recall that $w>0$, so this amounts to choosing the pairs with highest cosine similarity.
\section{Active learning experiments}
\subsection{Experimental details}
\label{sec:experimental-details}

We simulate active learning with the imputed labels so that we can compare different algorithms without performing expensive gold label collection for each algorithm.
We collect $n_1 = 2048$ examples in the seed set,
and use $k=10$ rounds of active learning for QQP and $k=4$ for WikiQA,
as WikiQA is much smaller.
At round $i$, we query $n_i = n_1 \cdot (3/2)^{i-1}$ new labels.
The exponentially growing $n_i$ helps us avoid wasting queries in early rounds, when the model is worse,
and also makes training faster in the early rounds.
These choices imply a total labeling budget $n$ of 232,100 for QQP and 16,640 for WikiQA.
For both datasets, $n$ is slightly less than $|\Dtrainin|$ (257,421 for QQP and 20,360 for WikiQA),
thus ensuring a meaningful comparison with training on heuristic data.
We retrieve $m=1000$ nearest neighbors per $x_1 \in \sX_1$ for QQP and $m=100$ for WikiQA.
We run all active learning experiments with three different random seeds and report the mean.
Training details are given in \refapp{app-experiments}.

\subsection{Main results}
\begin{table}[t]
\centering
\small
\begin{tabular}{lrrrr}
\toprule
\multirow{2}{*}{Method} & \multicolumn{2}{c}{QQP} & \multicolumn{2}{c}{WikiQA} \\
& P@R20 & \multicolumn{1}{c}{AP} & P@R20 & \multicolumn{1}{c}{AP} \\
\midrule
Random      &  $4.7\%$ & $ 2.8\%$ & $1.3\%$ & $1.3\%$ \\
Stated data & $29.3\%$ & $15.4\%$ & $0.7\%$ & $2.2\%$ \\
Static Ret. & $49.2\%$ & $25.1\%$ & $13.9\%$ & $8.2\%$ \\
\midrule
Adapt. Ret. & $59.1\%$ &  $32.4\%$ & $27.1\%$ & $15.1\%$ \\
Uncertainty & $\bf 60.2\%$ & $\bf 32.5\%$ & $\bf 32.4\%$ & $\bf 20.1\%$ \\
\bottomrule
\end{tabular}
\caption{
Main results comparing different data collection strategies on QQP and WikiQA.
The two active learning methods, adaptive retrieval and uncertainty sampling, greatly outperform other methods. }
\label{tab:main}
\end{table}

\begin{table}[t]
\centering
\small
\begin{tabular}{lrr}
\toprule
Positives Found     & \multicolumn{1}{c}{QQP} & \multicolumn{1}{c}{WikiQA} \\
\midrule
Random sampling     &       1 &      1 \\
Static retrieval    &  16,422 &    169 \\
Adaptive retrieval  & \textbf{103,181} &   \textbf{757} \\
Uncertainty sampling   &  87,594 &   742  \\
\midrule                      
Total examples collected  & 232,100 & 16,640 \\
\bottomrule
\end{tabular}
\caption{
Number of positive points collected by different methods.
All methods collect the same number of total examples (last row).
}
\label{tab:positives}
\end{table}

We now compare the two active learning methods, adaptive retrieval and uncertainty sampling,
with training on $\Dtrainin$ and two other baselines.
Random sampling queries $n$ pairs uniformly at random,
which creates a very imbalanced dataset.
Static retrieval
queries the $n$ most similar pairs using the pre-trained BERT embedding, similar to \refsec{seed-set}.
\reftab{main} shows all-pairs evaluation for \ourmodel trained on these datasets.
The two active learning methods greatly outperform other methods:
Uncertainty sampling gets $32.5\%$ AP on QQP and $20.1\%$ on WikiQA,
while the best static data collection method, static retrieval,
gets only $25.1\%$ AP on QQP and $8.2\%$ AP on WikiQA.
Recall from \reftab{std-eval} that \vanillabert only achieved $2.4\%$ AP on both QQP and WikiQA.
When trained on the same data as \vanillabert, \ourmodel achieves much higher AP on QQP ($15.4\%$) but slightly lower AP on WikiQA ($2.2\%$).
Uncertainty sampling slightly outperforms adaptive retrieval on both datasets.

Achieving high precision across all pairs requires 
collecting both enough positive examples and useful negative examples.
Compared to random sampling and static retrieval,
active learning collects many more positive examples,
as shown in \reftab{positives}.
$\Dtrainin$ contains all positive examples, but models trained on it still have low AP on all pairs.
We conclude that the negative examples in $\Dtrainin$ are insufficient for generalization to all pairs,
while active learning chooses more useful negatives.

\subsection{Manual verification of imputed negatives}
\label{sec:manual-2}
As in \refsec{manual}, we manually labeled putative QQP false positives at the threshold where recall is $20\%$ for \ourmodel trained on either stated data or uncertainty sampling data.
For each, we labeled 50 putative false positives from $\Ddevnear$, and all putative false positives from $\Ddevrandom$ (12 for stated data, 0 for uncertainty sampling).

\paragraph{\ourmodel trained on $\Dtrainin$.}
$67\%$ (8 of 12) of the putative false positives on $\Ddevrandom$ were actual errors by the model, but only $36\%$ of putative false positives on $\Ddevnear$ were errors. 
Extrapolating from these results, we update our estimate of development set precision at $20\%$ recall from $28.4\%$ to $41.4\%$.

Overall, this model makes some more reasonable mistakes than the \vanillabert model, though its precision is still not that high.

\paragraph{\ourmodel model with uncertainty sampling.}
Only $32\%$ of putative false positives from $\Ddevnear$ were real errors, 
significantly less than the $72\%$ for \vanillabert trained on $\Dtrainin$
($p = 7 \times 10^{-5}$, Mann-Whitney U test).
Extrapolating from these results, we update our estimate of development set precision at $20\%$ recall from $55.1\%$ to $79.3\%$,
showing that uncertainty sampling yields a more precise model than our imputed labels indicate. 
In fact, this model provides a high-precision way to identify paraphrase pairs not annotated in the original dataset.

\subsection{Comparison with stratified sampling}
\begin{table}[t]
\centering
\small
\begin{tabular}{lrrrr}
\toprule
\multirow{2}{*}{Method} & \multicolumn{2}{c}{QQP} & \multicolumn{2}{c}{WikiQA} \\
& P@R20 & \multicolumn{1}{c}{AP} & P@R20 & \multicolumn{1}{c}{AP} \\
\midrule
Strat. match    & $35.1\%$ & $19.0\%$ & $13.2\%$ & $8.5\%$ \\
Strat. all pos. & $41.6\%$ & $22.2\%$ & $13.7\%$ & $9.1\%$ \\
\midrule
Adapt. Ret.       & $59.1\%$ &  $32.4\%$ & $27.1\%$ & $15.1\%$ \\
Uncertainty       & $\bf 60.2\%$ & $\bf 32.5\%$ & $\bf 32.4\%$ & $\bf 20.1\%$ \\
\bottomrule
\end{tabular}
\caption{
Even though stratified sampling has access to oracle information,
active learning performs better by collecting more informative negative examples.
}
\label{tab:stratified}
\end{table}

Next, we further confirm that having all the positive examples is not sufficient for high precision.
In \reftab{stratified}, we compare with two variants of stratified sampling, in which positive and negative examples are independently subsampled at a desired ratio \citep{attenberg2010why}.
First, we randomly sample positive and negative training examples to match the number of positives and negatives collected by uncertainty sampling, the best active learning method for both datasets (``Strat. match'' in \reftab{stratified}).
Second, we trained on all positive examples and added negatives to match the number of positives on QQP or match the active learning total budget on WikiQA (``Strat. all pos.'').\footnote{This aligns better with the original WikiQA dataset, which has many more negatives than positives.}
For QQP, this yielded a slightly larger dataset than the first setting.
Note that stratified sampling requires oracle 
information: It assumes the ability to sample uniformly from all positives, even though this set is not known before data collection begins.
Nonetheless, stratified sampling trails uncertainty sampling by more than $10$ AP points on both datasets.
Since stratified sampling has access to all positives, active learning must be choosing more informative negative examples.

\subsection{Training other models on collected data}
\label{sec:app-transformer}
\begin{table}[t]
\centering
\small
\begin{tabular}{lcccc}
\toprule
\multirow{2}{*}{QQP Data} & \multicolumn{2}{c}{\ourmodel} & \multicolumn{2}{c}{\vanillabert}  \\
& P@R20 & AP & P@R20 & AP \\
\midrule
Stated data    & $29.3\%$ & $15.4\%$ &  $3.0\%$ & $2.4\%$ \\
Static Ret.    & $49.2\%$ & $25.1\%$ &  $4.6\%$ & $1.9\%$ \\
Stratified     & $35.1\%$ & $19.0\%$ &  $29.0\%$ & $16.4\%$ \\
Uncertainty    & $\bf 60.2\%$ & $\bf 32.5\%$ & $23.6\%$ & $8.9\%$ \\
\bottomrule
\end{tabular}
\caption{Comparison on QQP of \ourmodel with \vanillabert.
Data collected by active learning (using \ourmodel)
is more useful for training \vanillabert than stated data or static retrieval data.
Stratified sampling here matches the label balance of the uncertainty sampling data.
}
\label{tab:transformer}
\end{table}

For QQP, data collected with active learning and \ourmodel is useful for training other models on the same task.
\reftab{transformer} shows that \vanillabert does better on data collected by active learning---using \ourmodel---compared to the original dataset or static retrieval.
\vanillabert performs best with stratified sampling; recall that this is not a comparable data collection strategy in our setting, as it requires oracle knowledge.
\ourmodel outperforms \vanillabert in all training conditions; we hypothesize that the cosine similarity structure helps it generalize more robustly to pairs of unrelated questions.
However, \ourmodel trained on stated data does not do as well on WikiQA,
as shown in \reftab{main}.

\subsection{Data efficiency}
Adaptivity is crucial for getting high AP with less labeled data.
Static retrieval, the best static data collection method, gets $21.9\%$ dev AP on QQP with the full budget of 232,100 examples.
Uncertainty sampling achieves a higher dev AP of $22.6\%$ after collecting only 16,640 examples,
for a $14\times$ data efficiency improvement.
See \refapp{app-learning-curve} for further analysis.

\subsection{Effect of seed set}
\begin{figure}[t]
\centering
\begin{subfigure}{0.5\linewidth}
  \centering
  \includegraphics[width=\linewidth]{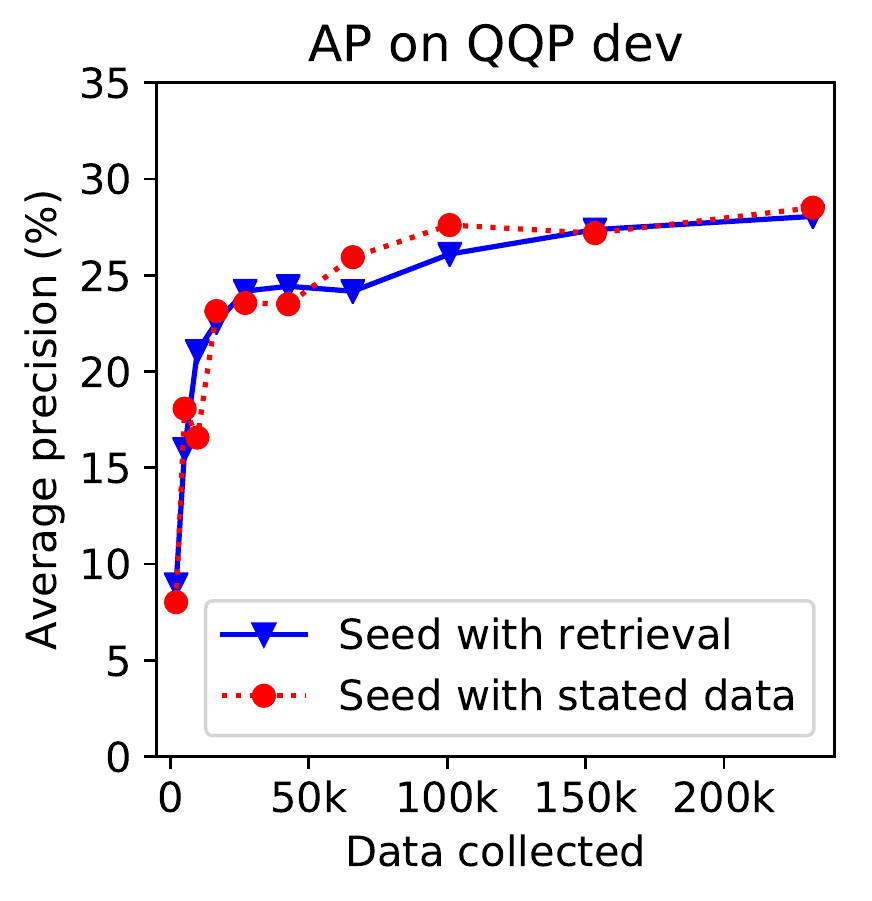}
  \caption{Average precision}
  \label{fig:stated-seed-ap}
\end{subfigure}%
\begin{subfigure}{0.5\linewidth}
  \centering
  \includegraphics[width=\linewidth]{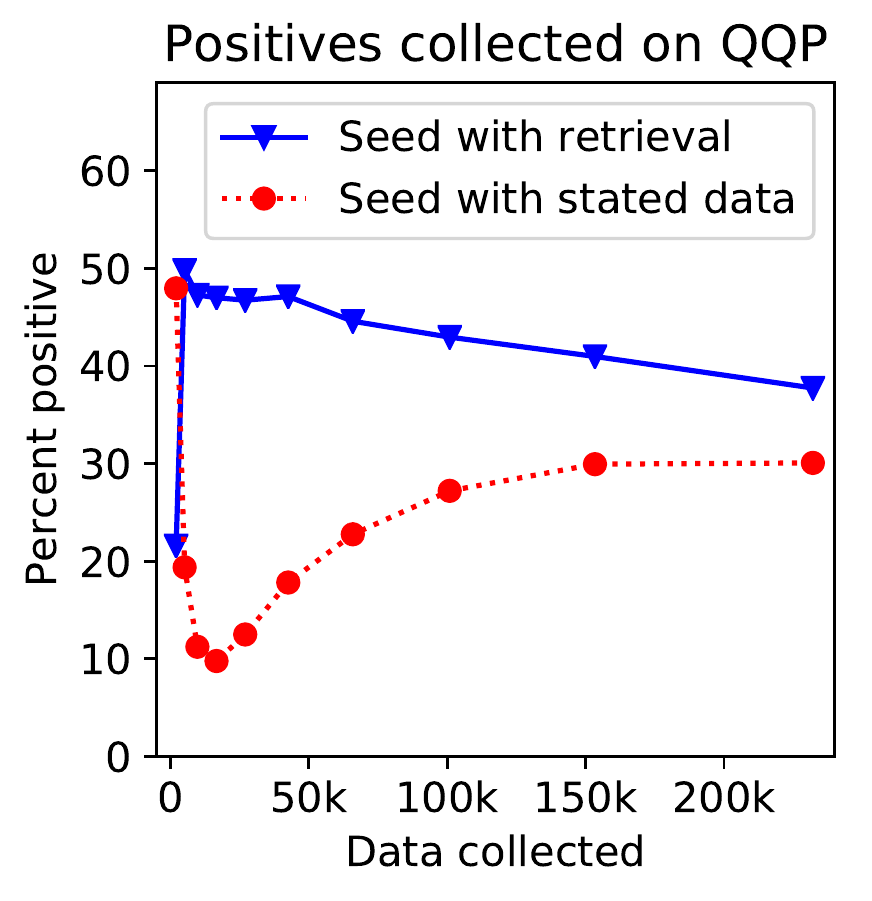}
  \caption{Positives collected}
  \label{fig:stated-seed-positives}
\end{subfigure}
\caption{
Uncertainty sampling on QQP using different seed sets.
(a) Seeding with stated data (one run) does similarly to seeding with retrieval (mean over three runs).
(b) Seeding with stated data makes the model poorly calibrated---points it is uncertain about are initially very unlikely to be positive. However, over time the model corrects this behavior.
} 
\label{fig:stated-seed}
\end{figure}

Our method is robust to choice of the initial seed set for uncertainty sampling.
We consider using stated data as the seed set,
instead of data chosen via static retrieval. 
As shown in \reffig{stated-seed},
seeding with stated data performs about as well as static retrieval in terms of AP.
Since the stated data artificially overrepresents positive examples, 
the model trained on stated data is initially miscalibrated---the points it is uncertain about are actually almost all negative points.
Therefore, uncertainty sampling initially collects very few additional positive examples.
Over time, adaptively querying new data helps correct for this bias.

\section{Discussion and related work} 
In this paper, we have studied how to collect training data that enables generalization to extremely imbalanced test data in pairwise tasks.
State-of-the-art models trained on standard, heuristically collected datasets have very low average precision when evaluated on imbalanced test data,
while active learning leads to much better average precision.

\citet{gillick2019learning} propose a similar model for entity linking and mine hard negative examples, an approach related to adaptive retrieval.
However, they have abundant labeled data, whereas we study data collection with a limited labeling budget. 

Work in information retrieval often attempts to maximize precision across all pairs of test objects.
Machine learning models are commonly used to  re-rank candidate pairs from an upstream retriever \citep{chen2017reading,nogueira2019passage},
while our method learns embeddings to improve the initial retrieval step.
Distant supervision has been used to train end-to-end retrieval models for question answering \citep{lee2019latent},
but does not extend to other tasks like paraphrase detection.
Other work on duplicate question detection on community QA forums
trains on labels generated by forum users \citep{dossantos2015hybrid}.
\citet{hoogeveen2016cqadupstack} show that these datasets tend to have many false negatives
and suggests additional labeling to correct this problem;
active learning provides one way to choose informative pairs to label.

Extreme label imbalance is an important challenge in many non-pairwise NLP tasks,
including
document classification \citep{lewis2004rcv1} and
relation extraction \citep{zhang2017tacred}.
Most prior work focuses on sampling a fixed training dataset 
\citep{chawla2004imbalanced,sun2009strategies,dendamrongvit2009undersampling},
whereas our work explores data collection.
\citet{attenberg2010why} find stratified sampling outperforms active learning in non-pairwise imbalanced tasks, primarily due to the difficulty of finding a useful seed set.
We find pre-trained embeddings effective for seed set collection in pairwise tasks.

\citet{zhang2019selection} found that the frequency of questions in QQP leaks information about the label.
Evaluating on all pairs avoids such artifacts, as every test utterance appears in the same number of examples.
\citet{zhang2019selection} re-weight the original dataset to avoid these biases,
but re-weighting cannot compensate for the absence of some types of negative examples,
unlike active learning.

Many pairwise datasets are generated by asking crowdworkers to generate part or all of the input $x$ \citep{bowman2015large,mostafazadeh2016corpus}.
Having crowdworkers generate text increases the risk of introducing artifacts
\citep{schwartz2017roc,poliak2018hypothesis},
while our pool-based approach considers the entire distribution of utterance pairs.

We use active learning, specifically uncertainty sampling \cite{lewis1994sequential}, 
to create a balanced training set that leads to models that generalize to the full imbalanced distribution. \citet{ertekin2007learning} argues that active learning is capable of providing balanced classes to the learning algorithm by selecting examples close to the decision boundary. Furthermore, active learning can generalize to the full distribution, both empirically \citep{settles2009active,yang2018benchmark} and theoretically \citep{balcan2007margin,balcan2013active,mussmann2018uncertainty}.

Finally, this paper addresses two central concerns in NLP today: How to construct fair but challenging tests of generalization \citep{geiger2019posing},
and how to collect training data in a way that improves generalization.
Evaluating on extremely imbalanced all-pairs data has several advantages over other tests of generalization. 
Our examples are realistic and natural, unlike adversarial perturbations \citep{ebrahimi2018hotflip,alzantot2018adversarial}, and diverse, unlike hand-crafted tests of specific phenomena \citep{glockner2018breaking,naik2018stress,mccoy2019right}. 
Since we allow querying the label of any training example,
generalization to our test data is achievable, while out-of-domain generalization \citep{levy2017zero,yogatama2019learning,talmor2019generalization} may be statistically impossible. 
Our work thus offers a natural, challenging, and practically relevant testbed to study both generalization and data collection.

\paragraph{Reproducibility.} Code and data needed to reproduce all results can be found on the CodaLab platform at \url{https://bit.ly/2GzJAgM}.

\section*{Acknowledgments}
This work was supported by a PECASE Award, NSF Award Grant no. 1805310, and NSF Graduate Fellowship DGE-1656518.
We thank Chris Manning, Michael Xie, Dallas Card, and other members of the Stanford NLP Group for their helpful comments.

\bibliography{bibliography}
\bibliographystyle{acl_natbib}

\appendix

\section{Experimental details}
\label{sec:app-experiments}

\subsection{Training details}
\label{sec:app-training}
At each round of active learning, we train for $2$ epochs.
We train without dropout, as dropout artificially lowers cosine similarities at training time.
We apply batch normalization \citep{ioffe2015batch} to the cosine similarity layer to rescale the cosine similarities, as they often are very close to 1.
We initialize $w$ and $b$ so that high cosine similarities correspond to the positive label,
and constrain $w$ to be nonnegative during training.
We use a maximum sequence length of 128 word piece tokens.
To compensate for BERT's low learning rate, we increased the learning rate on the $w$ and $b$ parameters by a factor of $10^4$.

Below in \reftab{training-details}, we show hyperparameters for training.
Hyperparameters were tuned on the development set of QQP; 
we found these same hyperparameters also worked well for WikiQA, and so we did not tune them separately for WikiQA.
In most cases, we used the default hyperparameters for BERT.

\begin{table}[th]
\centering
\begin{tabular}{lc}
\toprule
Hyperparameter & Value \\
\midrule
Learning rate & $2 \times 10^{-5}$ \\
Training epochs & 2 \\
Weight decay & $0$ \\
Optimizer & AdamW \\
AdamW Epsilon & $1 \times 10^{-6}$ \\
Batch size & 16 \\
\bottomrule
\end{tabular}
\caption{Hyperparameter choices for QQP and WikiQA}
\label{tab:training-details}
\end{table}

At the end of training, we freeze the embeddings $e_\theta$ and train the output layer parameters $w$ and $b$ to convergence,
to improve uncertainty estimates for uncertainty sampling.
This process amounts to training a two-parameter logistic regression model. 
We optimize this using (batch) gradient descent with learning rate $1$ and $10,000$ iterations. 
When training this model, we normalize the cosine similarity feature to have zero mean and unit variance across the training dataset.
Training this was very fast compared to running the embedding model.

Each experiment was conducted with a single GPU, most commonly a TITAN V or TITAN Xp.
Running one complete uncertainty sampling experiment (i.e., 10 rounds of data collection and model training for QQP, 4 for WikiQA) on a machine with one TITAN V GPU
takes about 9 hours for QQP and about 30 minutes for WikiQA.
Recall that \ourmodel only adds two additional parameters, $w$ and $b$, on top of a BERT model;
we use the uncased BERT-base pre-trained model which has 110M parameters.

\subsection{Evaluation details}
\label{sec:app-evaluation}
To evaluate a given scoring function $S$ at threshold $\gamma$ on a test set $\Dtestpairs$, 
we must compute the number of true positives $\TP(S, \gamma)$, false positives $\FP(S, \gamma)$, and false negatives $\FN(S, \gamma)$.
True positives and false negatives are computationally easy to compute, as they only require evaluating $S(x)$ on all the positive inputs $x$ in $\Dtestpairs$.
However, without any structural assumptions on $S$, it is computationally infeasible to exactly compute the number of false positives, as that would require evaluating $S$ on every negative example in $\Dtestpairs$, which is too large to enumerate.

Therefore, we devise an approach to compute an unbiased, low-variance estimate of $\FP(S, \gamma)$.
Recall that this term is defined as 
\begin{align}
\FP(S,\gamma) &= \sum_{x \in \Dtestpairs} \textbf{1}[ y(x)=0  \wedge S(x) > \gamma ] 
            \\&= \sum_{x \in \Dtestneg} \textbf{1}[ S(x) > \gamma ]
\end{align}
where $\Dtestneg$ denotes the set of all negative examples in $\Dtestpairs$.

One approach to estimating $\FP(S, \gamma)$ would be simply to randomly subsample $\Dtestneg$ to some smaller set $R$, count the number of false positives in $R$, and then multiply the count by $|\Dtestneg| / |R|$.
This would be an unbiased estimate of $\Dtestneg$, but has high variance when the rate of false positive errors is low.
For example, if $|\Dtestneg| = 10^{10}$, $|R| = 10^6$, and the model makes a false positive error on $1$ in $10^6$ examples in $\Dtestneg$, then $\FP(S, \gamma) = 10^4$.
However, with probability \[
\left(1 - \frac1{10^6}\right)^{10^6} \approx 1/e \approx 0.368,
\]
$R$ will contain no false positives, so we will estimate $\FP(S, \gamma)$ as $0$.
A similar calculation shows that the probability of having exactly one false positive in $R$ is also roughly $1/e$, which means that with probability roughly $1 - 2/e \approx 0.264$, we will have at least two false positives in $R$, and therefore overestimate $\FP(S, \gamma)$ by at least a factor of two.

To get a lower variance estimate $\hatFP(S, \gamma)$, we preferentially sample from likely false positives and use importance weighting to get an unbiased estimate of $\FP(S, \gamma)$.
In particular, we construct $\Dtestnear$ to be the pairs in $\Dtestpairs$ with nearby pre-trained BERT embeddings, analogously to how we create the seed set in \refsec{seed-set}.
Points with nearby BERT embeddings are likely to look similar are therefore more likely to be false positives.
Note that 
\begin{align}
\FP(S, \gamma) &= \sum_{x \in \Dtestnear} \textbf{1}[ S(x) > \gamma]  \nonumber
\\&+ \sum_{x \in \Dtestneg \setminus \Dtestnear} \textbf{1}[ S(x) > \gamma]
\\&= \sum_{x \in \Dtestnear} \textbf{1}[ S(x) > \gamma]  \nonumber
\\&+ \wrandom \cdot \mathbb{E}_{x \sim \text{Unif}(\Dtestneg \setminus \Dtestnear)} \textbf{1}[S(x) > \gamma],
\end{align}
where we define $\wrandom = |\Dtestneg|  - |\Dtestnear|$.

We can compute the first term exactly, since $\Dtestnear$ is small enough to enumerate, and approximate the second term as 
\begin{align}
\wrandom \cdot \frac1{|\Dtestrandom|} \sum_{x \in \Dtestrandom} \textbf{1}[S(x) > \gamma],
\end{align}
where $\Dtestrandom$ is a uniformly random subset of $\Dtestneg \setminus \Dtestnear$.
Therefore, our final estimate is
\begin{align}
\hat{\FP}(S, \gamma) &=  \sum_{x \in \Dtestnear} \textbf{1}[ S(x) > \gamma]  \nonumber
\\&+ \frac{\wrandom}{|\Dtestrandom|} \sum_{x \in \Dtestrandom} \textbf{1}[S(x) > \gamma].
\end{align}

\subsection{Incorporating manual labels}
\label{sec:app-manual-update}
In \refsec{manual}, we manually label examples that were automatically labeled as false positives,
and use this to improve our estimates of the true model precision.
We manually label randomly chosen putative false positives from both $\Ddevnear$ and $\Ddevrandom$, 
and use this to estimate
the proportion of putative false positives in each set that are real false positives.
Let $\hat{p}_{\text{near}}$ denote the estimated fraction of putative false positives in $\Ddevnear$ that are real false positives, and $\hat{p}_{\text{rand}}$ be the analogous quantity for $\Ddevrandom$.
Our updated estimate $\hat{\FP}_{\text{manual}}(S, \gamma)$ is then defined as
\begin{align}
\hat{\FP}_{\text{manual}}(S, \gamma) &=\hat{p}_{\text{near}} \sum_{x \in \Dtestnear} \textbf{1}[S(x) > \gamma] \nonumber
\\&+ \frac{\hat{p}_{\text{random}} \cdot \wrandom}{|\Dtestrandom|} \sum_{x \in \Dtestrandom} \textbf{1}[S(x) > \gamma].
\end{align}

We then compute precision using $\hat{\FP}_{\text{manual}}(S, \gamma)$ in place of $\hat{\FP}(S, \gamma)$.

\subsection{Comparison with GLUE QQP data}
\label{sec:app-glue}
For a few reasons, our QQP in-domain accuracy numbers are lower than those on the GLUE leaderboard, which has accuracies in the low $90$'s. 
First, our training set is smaller (257K examples versus 364K).
Second, our split is more challenging because the model does not see the same questions or even same paraphrase clusters at training time and test time.
Finally, our test set is more balanced ($58\%$ negative) than the GLUE QQP dev set ($63\%$ negative; test set balance is unknown).
As a sanity check, we confirmed that our RoBERTa implementation can achieve $91.5\%$ dev accuracy when trained and tested on the GLUE train/dev split, in line with previously reported results \citep{liu2019roberta}.


\section{Additional experimental results}

\subsection{Learning curves and data efficiency}
\label{sec:app-learning-curve}

\begin{figure}[h]
\centering
\small
\begin{subfigure}{0.5\linewidth}
  \centering
  \includegraphics[width=\linewidth]{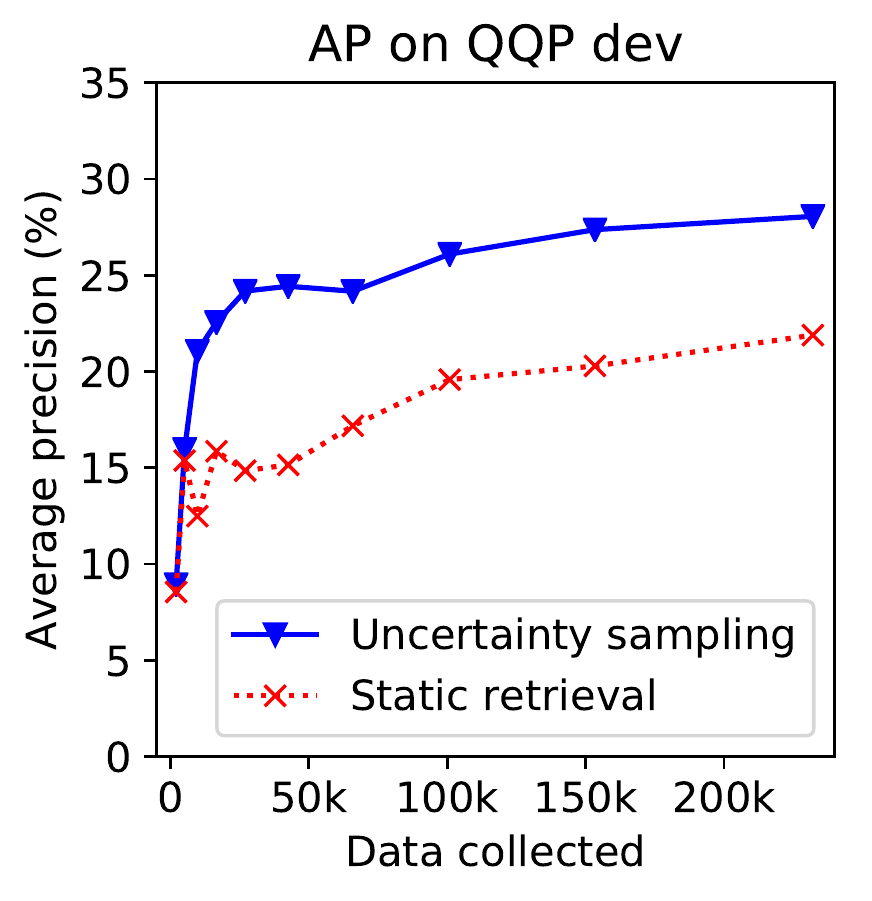}
  \caption{Average precision}
  \label{fig:learning-curve}
\end{subfigure}%
\begin{subfigure}{0.5\linewidth}
  \centering
  \includegraphics[width=\linewidth]{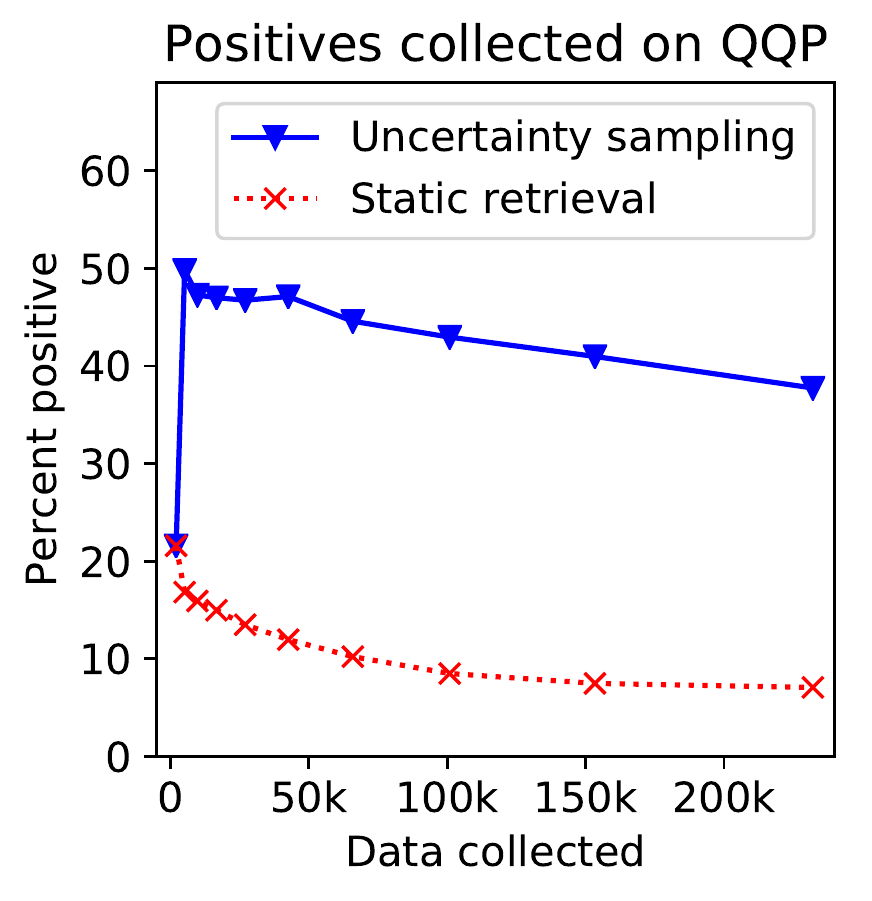}
  \caption{Positives collected}
  \label{fig:positives-curve}
\end{subfigure}
\caption{
Uncertainty sampling compared with matching amounts of static retrieval data on QQP.
(a) Average precision is higher for uncertainty sampling.
(b) Percent of all collected data that is positive. Adaptivity helps uncertainty sampling collect more positives.
}
\label{fig:curves}
\end{figure}

In \reffig{learning-curve}, we plot average precision on the QQP dev set
for our model after each round of uncertainty sampling.
For comparison, we show a model trained on the same amount of data collected via static retrieval, the best-performing static data collection method.
Uncertainty sampling leads to higher AP with much less data.
For example,
uncertainty sampling only needs to collect 16,640 examples
to surpass the average precision of static retrieval collecting all 232,100 examples, for a $14\times$ data efficiency improvement.
A big factor for the success of uncertainty sampling is its ability to collect many more positive examples than static retrieval, as shown in 
\reffig{positives-curve}. 
Static retrieval collects fewer positives over time, as it exhausts the set of positives that are easy to identify.
However, uncertainty sampling collects many more positives, especially after the first round of training, because it improves its embeddings over time.

\end{document}